\documentclass[runningheads]{llncs}

\usepackage[year=2024,ID=4811]{eccv}
\usepackage{colortbl}
\usepackage{todonotes}
\usepackage{gensymb}

\usepackage{multicol}
\usepackage{floatrow}
\usepackage{multirow}
\usepackage{subcaption}
\usepackage{array}
\newcolumntype{H}{>{\setbox0=\hbox\bgroup}c<{\egroup}@{}}
\usepackage{eccvabbrv}

\usepackage{graphicx}
\usepackage{booktabs}

\usepackage[accsupp]{axessibility}  % Improves PDF readability for those with disabilities.

\usepackage[pagebackref,breaklinks,colorlinks]{hyperref}
\usepackage{orcidlink}
\definecolor{tabfirst}{rgb}{1, 0.7, 0.7} % red
\definecolor{tabsecond}{rgb}{1, 0.85, 0.7} % orange
\definecolor{tabthird}{rgb}{1, 1, 0.7} % yellow
\definecolor{lightgray}{rgb}{0.9, 0.9, 0.9} % gray

\begin{document}

\title{Mamba-ND: Selective State Space Modeling for Multi-Dimensional Data} 

\titlerunning{Mamba-ND}

\author{Shufan Li, Harkanwar Singh, Aditya Grover \\
\{jacklishufan,harkanwarsingh,adityag\}@cs.ucla.edu \\}
 \institute{
University of California, Los Angeles}

\authorrunning{Li et al.}

% % TODO FINAL: Replace with your institution list.
% Springer Heidelberg, Tiergartenstr.~17, 69121 Heidelberg, Germany

\def\ourmethod{Mamba-ND}

\maketitle

\begin{abstract}
In recent years, Transformers have become the de-facto architecture for sequence modeling on text and multi-dimensional data, such as images and video.
However, the use of self-attention layers in a Transformer incurs prohibitive compute and memory complexity that scales quadratically w.r.t. the sequence length. A recent architecture, Mamba, based on state space models has been shown to achieve comparable performance for modeling text sequences, while scaling linearly with the sequence length.
In this work, we present Mamba-ND, a generalized design extending the Mamba architecture to arbitrary multi-dimensional data. 
Our design alternatively unravels the input data across different dimensions following row-major orderings. 
We provide a systematic comparison of Mamba-ND with several other alternatives, based on prior multi-dimensional extensions such as Bi-directional LSTMs and S4ND. 
Empirically, we show that  Mamba-ND demonstrates performance competitive with the state-of-the-art on various multi-dimensional benchmarks, including ImageNet-1K classification, HMDB-51 and UCF-101 action recognition, ERA5 weather forecasting and BTCV 3D segmentation. Code is available at \href{https://github.com/jacklishufan/Mamba-ND}{https://github.com/jacklishufan/Mamba-ND}
% While Transformers have shown superior performance than ConvNets on a variety of multi-dimensional data, their self-attention layers exhibit quadratic complexity with respect to sequence length. This limits their scaling capabilities. Recently, selective state space models (sSSMs), Mamba in particular, exhibits stronger performance than transformers on 1-D sequences. Exploring whether such performances can be extended to higher dimensional data is an interesting topic. In this work, we conducted systematic comparison on various ways to extend sSSMs beyond 1D sequence. Many of these approaches draw inspiration from earlier works on sequencing modeling such as Bidirectional-LSTM. Using these results, we propose Mamba-ND, a multi-dimensional sSSM that extends Mamba to high dimensional data by processing each dimension in an alternating fashion. Mamba-ND demonstrate competitive performance on various tasks including ImageNet-1K classification, HMDB-51 action recognition, and ERA5 climate forecast. Code is available at https://github.com/jacklishufan/Mamba-ND

\keywords{State Space Models \and Multi-Dimensional Modeling }

\end{abstract}    
\section{Introduction}
\label{sec:intro}
\begin{figure}[h]
    \centering
    \includegraphics[width=0.8\columnwidth]{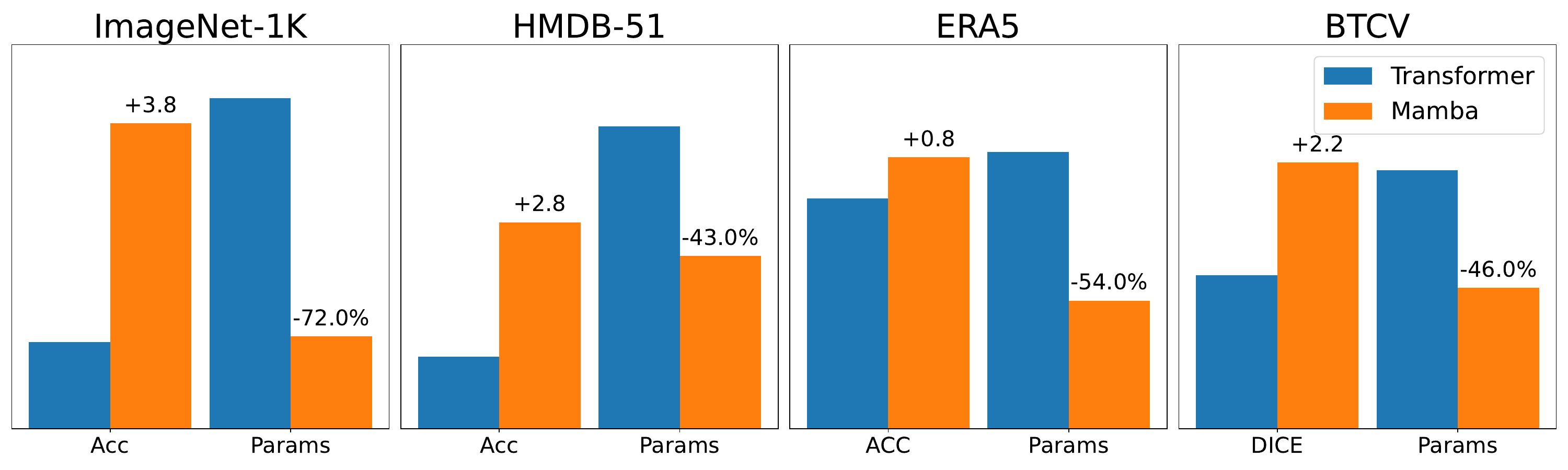}
    % \vspace{-2em}
    \caption{\textbf{\ourmethod~ outperforms Transformers while significantly reducing the number of parameters.} On ImageNet-1k, we compare against ViT \cite{vit}. On HMDB-51\cite{kuehne2011hmdb}, we compare against Video-Swin \cite{liu2022video}. On ERA5, we compare against Cli-ViT. On BTCV, we compare against UNETR\cite{hatamizadeh2022unetr}. \cite{nguyen2023climax}.
    }
    \label{fig:performance}
\end{figure}
The design of flexible and scalable neural network architectures is fundamental to the success of deep learning across diverse domains.
Convolutional neural networks~\cite{lecunHandwrittenDigitRecognition1989} excel at handling continuous data such as images, audio, and video. Recently, they have been surpassed by Transformers~\cite{vaswaniAttentionAllYou2017}, which process continuous data as a discrete sequence of patches~\cite{vit}. Despite their superior performance on many tasks, Transformer-based models struggle to scale to larger patch sequence as they scale quadratically with respect to sequence length. Most recently, a special kind of State Space Model (SSM) termed as Mamba~\cite{guMambaLinearTimeSequence2023}, has demonstrated stronger performance than Transformers while maintaining a linear complexity. However, the impressive performance of Mamba was shown on 1D text sequences. This leaves open the question whether Mamba can be effectively extended to multi-dimensional data such as images, video, or scientific datasets, which is the focus of this work.

% n whether SSM also hold an advantage against Transformers on high-dimensional data.

Unlike convolution or self-attention operations, which can be computed in parallel across the ND input data, Mamba requires a specific ordering of the data. Determining such an order is not an easy task. Building on past work in architecture design, we could consider many choices.
One naive approach is to flatten the data in row-major order. Intuitively, this is non-optimal because in this setting information only flows in one direction, which is suboptimal for multi-dimensional data with no default ordering.
Drawing inspiration from early works on Bi-directional LSTM, another alternative is to process the sequence in two directions at each layer and aggregate the results. We call this method Bi-SSM. While this design, in principle, allows information exchange between two arbitrary patches, two patches adjacent to each other spatially may have a huge distance between them on the computation graph. A natural extension of this method, which we call ND-SSM, is to process the input in $2D$ directions, where $D$ is the dimension of the data, and aggregate the results. 
Rather than picking orders in sequence, another possibility is to borrow inspiration from multi-head self-attention in transformers, wherein we can split the channels into multiple heads and let each head process an SSM in a different direction. This design is similar to ND-SSM but differs in that it has less computational burden per layer.

At a high level, there is also the question of block-level design, which specifies how Mamba layers are organized. For example, using the vanilla Mamba layer as a black box, one can still apply the bi-directional or N-directional design by processing the same input through multiple layers with different directions.

In this work, we conducted an extensive study on these possible design choices. Surprisingly, we find that simply alternating between three fixed row-major orderings is one of the best-performing strategies on both 2D and 3D data. Armed with these findings, we propose Mamba-ND, a surprisingly simple yet effective design to extend Mamba to multi-dimensional data. Mamba-ND does not introduce any complicated changes to 1D-SSM layers. By stacking 1D-SSM layers as black boxes and alternating the sequence order between each layer, Mamba-ND was able to surpass Transformer-based models on various tasks, including image classification, action recognition, and weather forecasting and 3D segmentation, with a lower parameter count. It also maintains a linear complexity with respect to input sequence length.

In summary, our main contributions are as follows. 
\begin{itemize}
    \item We propose \ourmethod~, which extends SSM to higher dimension through simply alternating sequence odering across layers.
    \item Compared with Transformers, our model is able to achieve stronger performance at much lower parameter count on a diverse set of tasks.
    % while considerably reducing the number of parameters.
    % \item Compared with Transformers, our model was able to achieve stronger performance at lower parameter count. Compared with ViT, our model achieves +1.5 increase in ImageNet-1K classification accuracy while reducing the number of parameters by 20.7\%. On HMDB-51 video classification task, \ourmethod~ our model achieves +0.9 increase compared with 3D Swin-Transformer while reducing the number of parameters by 39.0\%. On ERA5 5.625 degree 72 hour weather forecast tasks, \ourmethod~ achieves +0.7 gain in Anomaly Correlation Coefficient (ACC) while reducing the number of parameters by 44.5\% when compared with Cli-ViT. 
    \item We conduct extensive ablation studies on various more complicated approaches that extend SSM to multi-dimensional inputs. We find that complicated designs do not necessarily translate into stronger performance.  
    % These results should help future work on adapting 1D Sequence models to higher dimension.
\end{itemize}
\section{Related Works}
\label{sec:related}

\paragraph{Modeling 2D Data :}

Convolutional Neural Networks (CNNs) \cite{lecunHandwrittenDigitRecognition1989,resnet,huang2017densely,szegedy2015going,krizhevsky2012imagenet,simonyan2014very} have historically been the state-of-the-art approach for image recognition tasks. The receptive field of CNNs only grows linearly with the model depth. This limitation restricts the ability of CNNs to effectively capture global context.

More recently, Vision Transformers (ViTs) \cite{vit} have emerged as a strong candidate for vision tasks. These models first divide an input image into a grid of discrete patches and then process them as a 1D sequence. While they provide a global receptive field through self-attention mechanisms at every layer, such advantages come with the cost of quadratic computational and memory complexity with respect to input length, making them challenging to scale.

To address this limitation, several works have proposed hybrid architectures that incorporate attention mechanisms into CNNs \cite{wang2018non,chen2017sca,li2020attention} or introduce hierarchical designs into transformers \cite{liu2021swin,li2022mvitv2,dai2021coatnet,ding2022davit,dong2022cswin,lu2021container,tian2023integrally,wang2021pyramid,zhang2022hivit}. Other approaches, such as PixelRNN \cite{van2016pixel}, attempt to apply recurrent models, which are inherently designed for linear complexity and global context, to images. 
There has also been considerable efforts to apply state space models to 2D images, such as 2D-SSM\cite{baron20232} and S4ND \cite{s4nd}. Our work falls within this final category, wherein we specifically explore ways to adapt selective state space models to images.

\paragraph{Modeling 3D Data :}

Modeling videos has long presented a significant challenge. Earlier works \cite{carreira2017quo,tran2015learning} primarily focused on extending 2D ConvNets into the 3D domain. More recently, 3D Transformers \cite{akbari2021vatt,liu2022video,neimark2021video} have demonstrated superior performance. The unique characteristics of the temporal dimension in video data also inspired various video-specific high-performing designs. For example, some works leverage extracted optical flow features \cite{christoph2016spatiotemporal,feichtenhofer2016convolutional} or motion vectors \cite{zhang2016real,zhang2018real}. Others, like SlowFast \cite{feichtenhofer2019slowfast}, employ specific architectural designs to account for the fact that pixel values change less in the temporal dimension than in the spatial dimensions. In contrast to these approaches, our goal is to devise a generic framework for modeling multi-dimensional data. Consequently, \ourmethod~treats videos as simple 3D arrays of RGB pixels, making no additional assumptions about the temporal structure. This is crucial because in certain use cases such as weather forecasting, the third dimension of the data represents an additional spatial dimension instead of a temporal one.
% Some of these works add temporal attention on top of existing 2D Vision Transformers, while others attempt to apply self-attention in a pure 3D fashion.

S4ND \cite{s4nd} is a prior attempt to model videos using state space models (S4 \cite{s4}) and has achieved competitive performance. It directly extends the S4 formulation to higher dimensions and leverages the time invariance constraint to parallelize computations as the outer product of three 1D convolutions. In contrast to S4ND, we aim to extend selective state space models, i.e. Mamba, to higher dimensions, which do not adhere to linear time invariance (LTI). As a result, the convolutional trick is not applicable, necessitating alternative designs.

In addition to videos, we also consider the 3D climate forecast and medical image segmentation task. In both field transformers \cite{nguyen2023climax,bi2022pangu,hatamizadeh2022unetr,hatamizadeh2021swin} have achieved successes. However they face similar scaling challenges as in typical vision problems.

% \paragraph{Climate Modeling.}

% Climate modeling presents a significant challenge in the 3D domain. We specifically focus on global weather forecasting. Typically, this task is evaluated using the ERA5 reanalysis dataset \cite{hersbach2018era5}. Early works \cite{weber2020deep,weyn2021sub} employed CNN architectures. Pangu-Weather \cite{bi2022pangu} was the first to introduce Transformers to this domain. It adapted a 3D-Swin Transformer to handle atmospheric measurements in 3D spaces. ClimaX \cite{nguyen2023climax} extended the Vision Transformer to weather data and proposed an architecture called Cli-ViT. Using this architecture, it established a strong foundational model, called ClimaX, capable of various tasks such as forecasting and downscaling through large-scale pretraining on simulated data.

% Using state space models for climate modeling remains an unexplored area. We are the first to apply SSMs in the 3D weather forecasting task.

\section{Background}
\label{sec:background}

\subsection{State Space Models}
Recent state space models (SSMs) \cite{s4, s4nd} have shown superior performance on long sequences. Formally, SSMs model the input data using the following ordinary differential equation (ODE):
\begin{align}
h'(t) &= Ah(t) + Bx(t) \label{eq:1a} \\
y(t) &= Ch(t) + Dx(t) \label{eq:1b}
\end{align}
Here, $x(t)\in \mathbb{R}$ is a continuous input signal in the time domain, and $y(t)\in \mathbb{R}$ is a continuous output signal in the time domain. In modern SSMs, this ODE is approximated through discretization. One common discretization method is the zero-order hold (ZOH) rule, which gives the following difference equation:
\begin{align}
\bar{A} &= \exp(\Delta A) \label{eq:1c}\\
\bar{B} &= (\Delta A)^{-1} (\exp(\Delta A) - I) \cdot \Delta B \label{eq:1d} \\
h_t &= \bar{A}h_{t-1} + \bar{B}x_t \label{eq:1e}\\
y_t &= Ch_t \label{eq:1f}
\end{align}
Early works such as S4~\cite{s4} and S4ND~\cite{s4nd} assume linear time invariance (LTI). This constraint makes it possible to solve the above difference equation using a global convolution kernel. Selective state space models, i.e., Mamba \cite{guMambaLinearTimeSequence2023}, introduce time-varying parameters that do not follow the LTI assumption. In particular, $\Delta$, $B$, and $C$ become functions of the input signal $x(t)$. Consequently, from \cref{eq:1c,eq:1d,eq:1e,eq:1f}, $\bar{A}$ and $\bar{B}$ also become dependent on time. This makes it hard to parallelize the computation. A hardware-aware optimization that makes use of associative scan is employed to mitigate this issue.

\subsection{Mamba Layers}
\label{sec:ssm_layer}
Mamba \cite{guMambaLinearTimeSequence2023} proposed an implementation of selective state space model (sSSM) layer (\cref{fig:head} column 1). It consists of a 1D convolution, an SSM kernel, and a residual connection. In each layer, the input sequence is first processed by the convolution operation and then by the SSM kernel. The result is added back to the input through the residual connection. While the Conv1D operation can be easily extended to multiple dimensions with ConvND operations, it is non-trivial to convert the SSM to multiple dimensions.  Particularly, the convolution trick is not applicable since Mamba is not a time-invariant system. In this work, we discuss various alternative ways to extend SSM to higher dimensions by flattening the input data into 1D sequences.
\section{Methodology}
\label{sec:methodology}

\begin{figure*}[t]
    \centering
    \includegraphics[width=1.0\columnwidth]{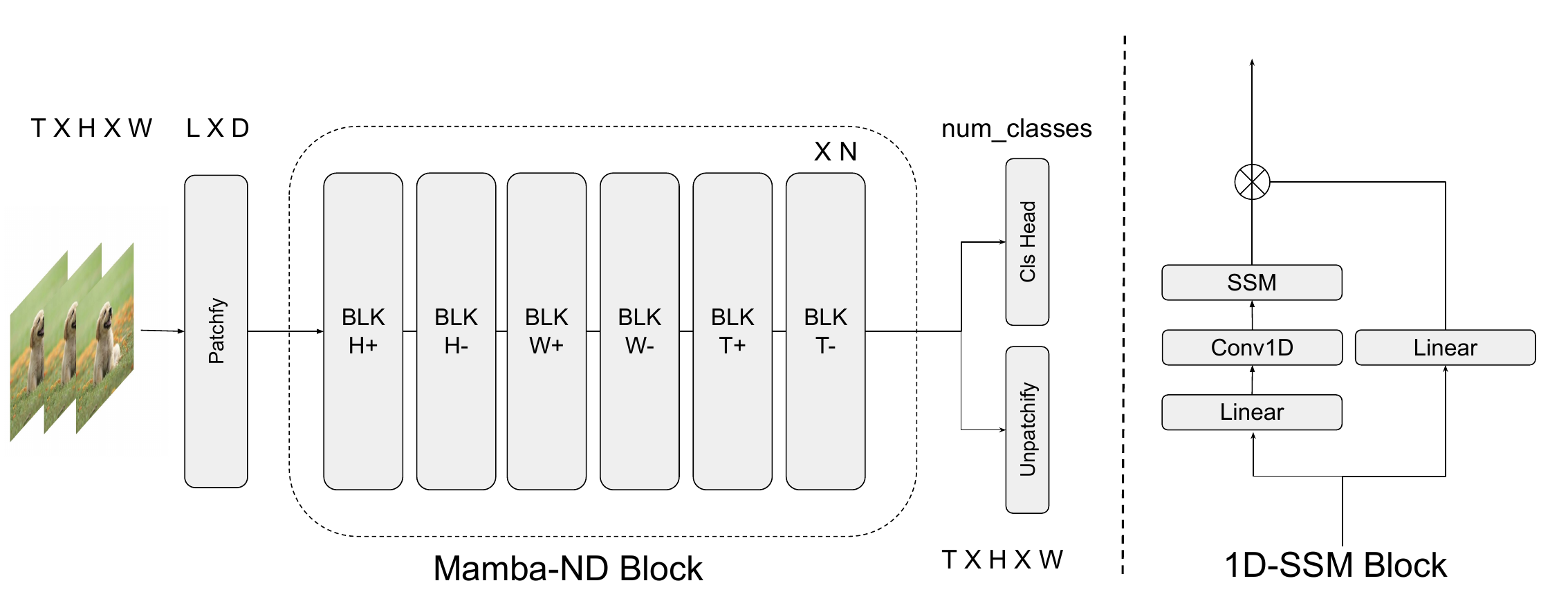}
    \vspace{-2em}
    \caption{\textbf{\ourmethod{} Architecture. } We visualize Mamba-3D as an example. Given 3D input, we patchify it into $L$ patches. During this process, we maintain the original 3D structure of the input. This sequence is then passed through $K$ Mamba-ND blocks, each of which consists of a chain of 1D Mamba layers that process the sequence in alternating orderings. In 3D space, we use the order H+H-W+W-T+T-. In 2D space, the sequence would be H+H-W+W-. Finally, the sequence is reshaped back to its original 3D structure and passed to task-specific heads for downstream processing. } 
    \label{fig:arch}
\end{figure*}

\begin{figure*}[h]
    \centering
    \includegraphics[width=1.0\columnwidth]{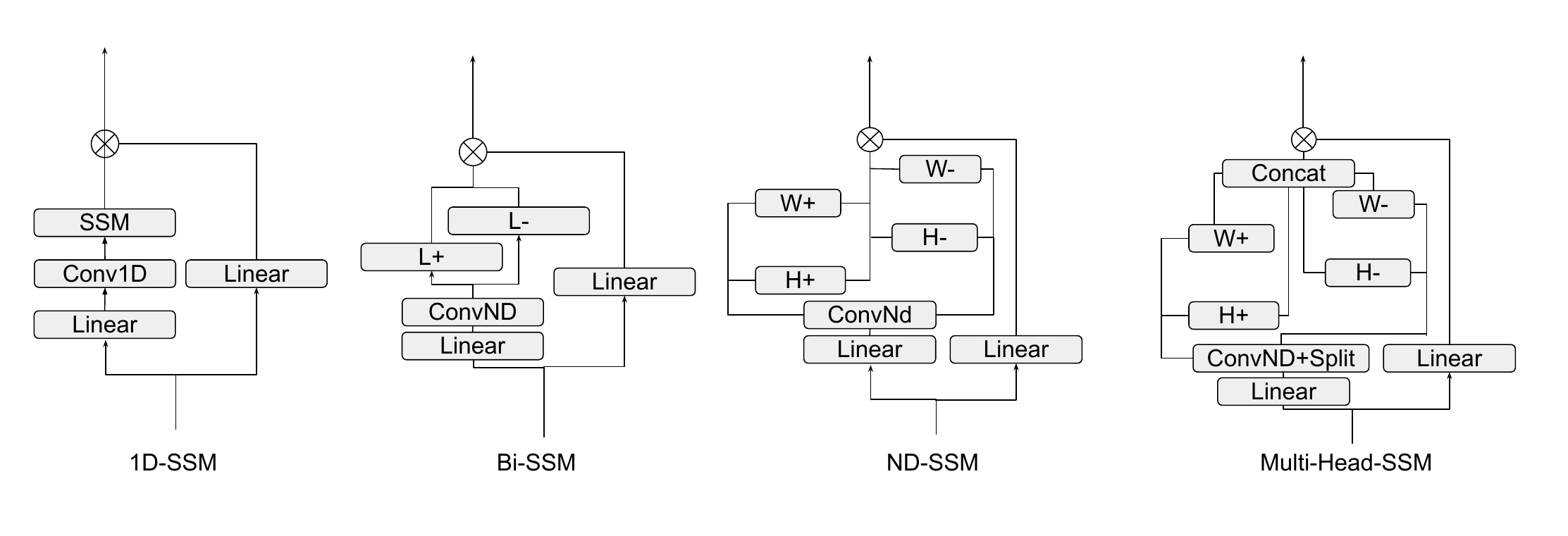}
    \vspace{-2em}
    \caption{\textbf{Variations of SSM Layer Design.} Col 1 represents the standard 1D SSM layer. Col 2 represents Bi-SSM, which adds bidirectionality in a similar fashion as LSTM. Col 3 represents ND-SSM block, which extends Bi-SSM to more directions. Col 4 represents multi-head SSM block inspired by multi-head attention in Transformers.
    }
    \label{fig:head}
\end{figure*}

We explores several approaches to adapt Mamba to multidimensional data. The key element of these designs is to devise a combination of sequence orderings to flatten the multidimensional data into 1D sequences. Intuitively, some level of bidirectional or multidirectional design is required to allow information exchange between two arbitrary data points in multidimensional space. This can be achieved at two levels: \textbf{layer level} and \textbf{block level}. As mentioned in \cref{sec:ssm_layer}, a Mamba layer consists of a 1D convolution, an SSM kernel, and a residual connection. One example of layer-level design is to pass the output of the convolution to two independent SSM kernels and sum up the results (\cref{fig:head}, col 2). Contrary to layer-level designs, block-level designs keep the internal design of a Mamba layer unchanged. Instead, it tries to achieve bidirectionality at the block level. For example, one such design is to alternate between each axis from one layer to another and apply a bidirectional design for each axis (\cref{fig:grouping}, row 2). One possible benefit of layer-level design is that it keeps the optimized fused kernel of a Mamba layer and the memory access pattern of the underlying array unchanged. 

\begin{figure*}[h]
    \centering
    \includegraphics[width=0.7\columnwidth]{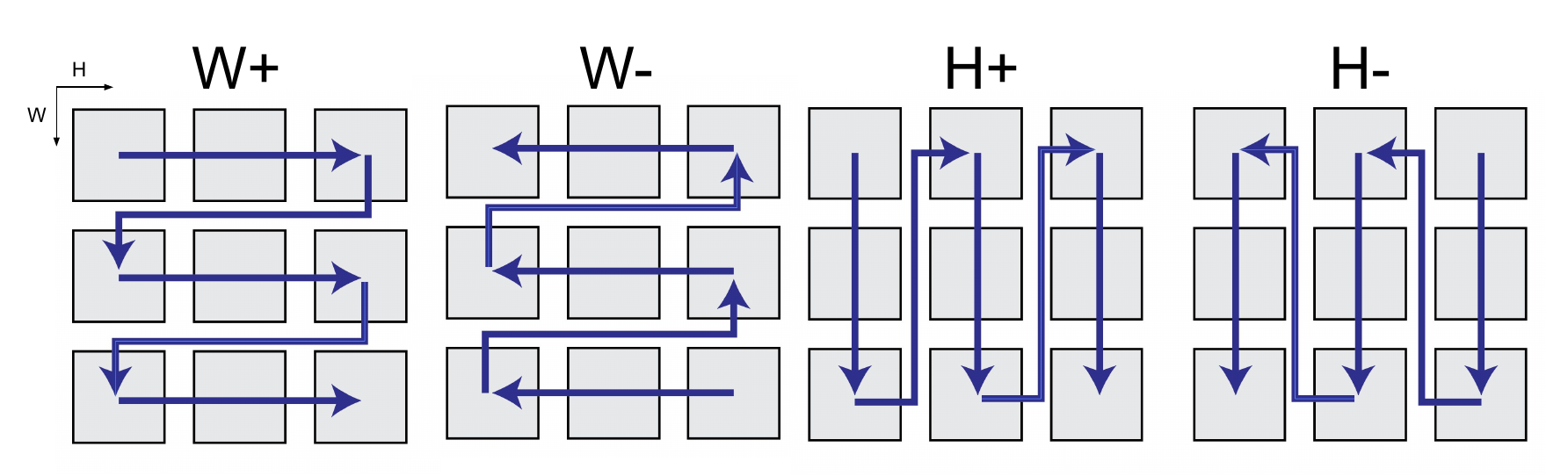}
    \vspace{0em}
    \caption{\textbf{Visualization of 2D scan orderings.} We visualize the set of possible all scan ordering on 2D data. Arrow indicates the scan order.
    }
    \label{fig:sc_2}
\end{figure*}

{
\begin{figure}[h]
\begin{subfigure}[b]{.45\linewidth}
% xx
% \begin{figure}[t]
    \centering
    \includegraphics[width=1.0\columnwidth]{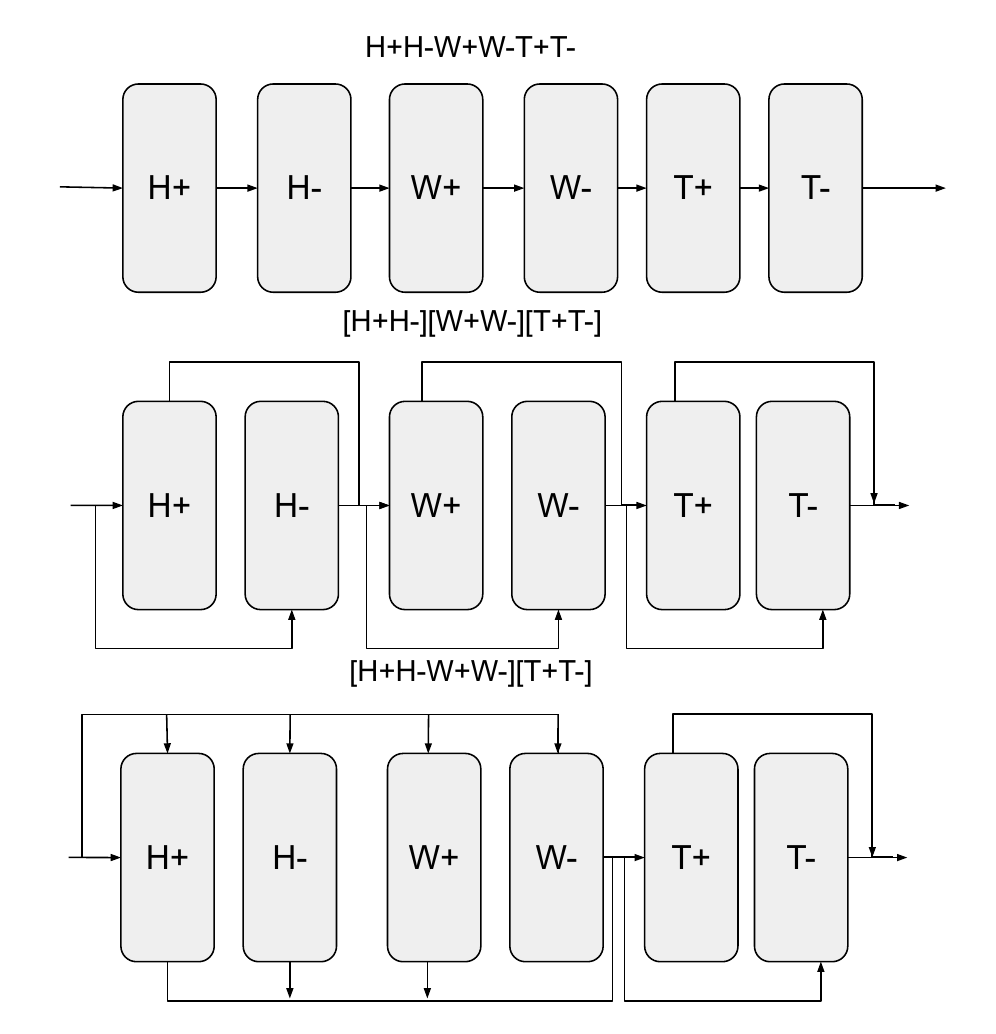}
    % \vspace{-2em}
    \caption{\textbf{Different ways of arranging Mamba layers.} The first row visualizes alternating-directional design. The second row visualizes bidirectional design. The third row visualizes quad-directional design.
    }
    \label{fig:grouping}
% \end{figure}
\end{subfigure}
% \begin{subfigure}[b]{.04\linewidth}
% \hphantom{1}
% \end{minipage}
\begin{subfigure}[b]{.45\linewidth}
% \begin{figure}[t]
    \centering
    \includegraphics[width=1.0\columnwidth]{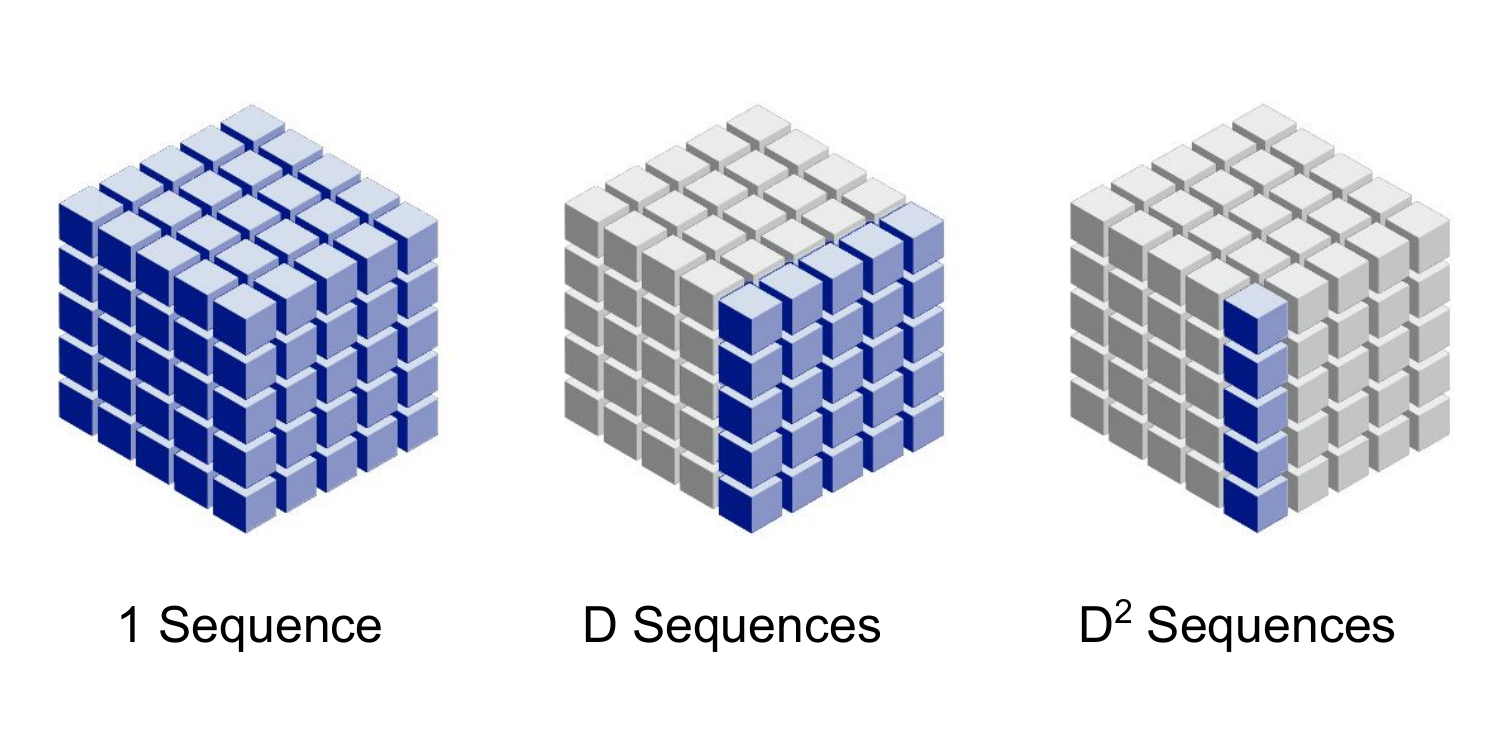}
    % \vspace{-2em}

    \caption{\textbf{Visualization of various Scan-Factorization policies.} Col 1: No factorization, there is only 1 sequence. Col 2: Factorizing the 3D sequence into $D$ 2D sequences, where $D$ is the length of a single dimension. Col 3: Factorizing the 3D sequence into $D^2$ 1D sequences. }

    \label{fig:factoring}
% \end{figure}
\end{subfigure}
\caption{\textbf{Visualization of block level design and factorization policies.}}
\end{figure}
}
\subsection{Scan Orderings}
Consider the input N-dimensional data $X$ of shape $D_1 \times D_2 \times \ldots \times D_N$, where $D_i$ is the length of data along the $i$th dimension. Let $L=\prod_{i=1}^N D_i$ be the total sequence length. There are a total of $L!$ possible ways of flattening $X$ into a 1D sequence. However, we only consider a subset of these choices, which we call the scan orderings. Formally, a scan ordering is obtained by permuting the order of axes of the original data $X$, and flattening it into a 1D sequence either in the forward or reverse direction. Since there are $N!$ permutations of $N$ axes and 2 possible directions, there are a total of $2N!$ scan orderings.
We denote a particular scan ordering $s$ as $(k_1, k_2, \ldots, k_N)\pm$, which represents the unique ordering obtained by first permuting the axis order from $1, 2, \ldots, N$ to $k_1, k_2, \ldots, k_N$, and then flattening the sequence in row-major order. The symbols $+$ and $-$ indicate whether the order of the final 1D sequence is in forward or reversed direction.

We focus on 2D data of shape $H \times W$ and 3D data of $T \times H \times W$. The 2D data has 4 possible orderings: $(HW)+$, $(HW)-$, $(WH)+$, and $(WH)-$. The 3D data has 12 possible orderings; examples include $(HWT)+$ and $(WHT)-$. To provide a concrete example, the ordering $(WH)-$ refers to first permuting the 2D data into a 2D array of $W \times H$, then flattening it into a 1D sequence in row-major order, and finally reversing the order of this 1D sequence.

For simplicity, we use $H$ to represent $WH$ and $W$ to represent $HW$ for 2D data. Similarly, we use $H$ to represent $TWH$, $W$ to represent $THW$, and $T$ for $HWT$ for 3D data. In this notation, $(WH)-$ becomes $H-$ and $(HWT)+$ becomes $T+$. We also use $L$ to represent $THW$ or $HW$, as this is the naive way of flattening 3D and 2D data to a 1D sequence without changing the memory layout. Notably, the last dimension will be traversed continuously. We visualize 2D scan orderings  in \cref{fig:sc_2}.

\subsection{Adapting the Mamba Layer}
\label{sec:head_dis}
We explore three alterations to the standard Mamba layer design, which are illustrated in \cref{fig:head}.

\noindent\textbf{Bi-SSM} layer passes the output of the convolution layer to two independent SSM kernels, one in the forward direction and another in the reversed direction. 

\noindent\textbf{ND-SSM} layer extends Bi-SSM by incorporating additional SSMs to accommodate other possible orderings. In the 2D case, there are four orderings $W+$, $W-$, $H+$, $H-$. 

\noindent\textbf{Multi-head SSM} layer is a mimic of the multi-head attention. It splits an input sequence of dimension $D$ into $H$ sequences of dimension $D/H$, where $H$ is the number of orderings. Each of the heads is then passed to separate SSM kernels in respective orderings. In the 2D case, the orderings are $W+$, $W-$, $H+$, $H-$. 

\subsection{Arranging Mamba Layers}
\label{sec:blk_dis}
In addition to making direct changes to the internal structure of Mamba, one can also change the way in which the layers are organized to achieve multi-directionality. We illustrate these variations in \cref{fig:grouping}.

\noindent\textbf{Alternating-Directional: H+H-W+W-T+T-} keeps the sequential ordering of Mamba layers and changes the direction of SSM in each layer in an alternating fashion. The ordering is H+H-W+W-T+T-

\noindent\textbf{Bi-Directional: [H+H-][W+W-][T+T-]} adopts a design on the block level. In each block, the input is passed to two Mamba layers at opposite directions. The ordering is [H+H-][W+W-][T+T-], where each $[.]$ denotes a bidirectional block consisting of two layers. To avoid confusion, we will explicitly refer to this method as [H+H-][W+W-][T+T-]. The term Bi-Directional will mostly be used for the Bi-SSM layer mentioned in \cref{sec:head_dis}. 

\noindent\textbf{Quad-Directional: [H+H-W+W-][T+T-]} builds on top of the [H+H-][W+W-][T+T-] design. It further groups the H and W directions. This design is inspired by works in video recognition e.g.,\cite{tran2018closer}, which factorize 3D convolution into a 2D operation on the spatial dimensions and a 1D operation in the temporal domain. 

There are more possible ways to organize multi-directional blocks, but they generally follow a similar design. Crucially, it is important to note that while each layer has a specific ordering, the SSM kernel operates on a single flattened input sequence. This means that all these layers have a global receptive field.

\subsection{Scan Factorization}
In order to mitigate the quadratic complexity of the Transformer, prior works \cite{ho2019axial} factorize full 3D attention into three 1D attentions along each axis. While SSMs already achieve linear complexity, the sequence length is still quadratic in the length of a single dimension. In this work, we also explore various ways of factorizing an SSM scan into multiple smaller scans. For an input array of dimensions $T \times H \times W$, the standard approach is to flatten it into a single sequence of length $THW$. Alternatively, we can factorize it into $T$ sequences with length $HW$, or $TH$ sequences with length $W$. We visualize these factorization techniques in \cref{fig:factoring}. Since SSM only retains one copy of a state per sequence in the linear scan process, increasing the number of sequences in parallel actually leads to an increase in memory consumption and training time because more hidden states need to be materialized in the GPU memory. However, we note that these sub-sequences need not be computed in parallel. Hence a better implementation in the future may reduce the memory cost of this design. Thus, we still investigate the performance of this design. 

\subsection{Final Design}

After extensive experiments, our final design uses the standard 1D-SSM layer and an alternating-directional: (H+H-W+W-T+T-) at the block level. We find that this simple design surprisingly outperforms more complicated ones. We provide more details of our experiments in \cref{sec:experiments}. 

Our overall image is shown in \cref{fig:arch}. Given multi-dimensional input data, we first patchify it into a 1D sequence. During this process, we keep track of the original 3D structure of the input. This sequence is then passed through $K$ Mamba-ND blocks, each of which consists of a chain of 1-D SSM layers that process the sequence in alternating orderings. In 2D space, we use the order H+H-W+W-. In 3D space, we use the order H+H-W+W-T+T-. Finally, the sequence is reshaped back to its original 3D structure and passed to task-specific heads for downstream tasks. 
\section{Experiments}
\label{sec:experiments}

\setlength{\arrayrulewidth}{0.8pt}

\paragraph{Datasets and Setups.}
We aim to evaluate the effectiveness of \ourmethod~ on various multi-dimensional data tasks. Specifically, we use ImageNet-1K \cite{dengImageNetLargescaleHierarchical2009} for image classification, HMDB-51 \cite{kuehne2011hmdb} and UCF-101 \cite{soomro2012ucf101} for action recognition, ERA5 5.625-degree for weather forecasting \cite{hersbach2018era5} and BTCV\cite{btcv} for 3D segmentation. ImageNet-1K is a large-scale dataset containing 1.2 million images across 1000 classes, HMDB-51 and UCF-101 are action recognition datasets comprising 7,030 and 13,320 video clips respectively, BTCV consists of abdominal CT scans of 30 subjects, among which 6 are selected as validation set. ERA5 consists of 3D atmospheric weather measurements, such as temperature and wind speed, across 13 pressure levels. We use the standard train and validation split for ImageNet, split1 of HMDB-51, and data from the years 1979-2016 for ERA5 as the training set, data from 2017 as the validation set, and data from 2018 as the test set. 

\paragraph{Metrics}
We measure top-1 accuracy for image classification and action recognition tasks. For weather forecasting, we report both the Residual Mean Squared Error (RMSE) and the Anomaly Correlation Coefficient (ACC). For 3D CT segmentation, we report the DICE score.
% , which is calculated as:

% \begin{equation}
% ACC = \frac{\sum (f - c)(a - c)}{\sqrt{\sum (f - c)^2 \sum (a - c)^2}}
% \end{equation}

% Here, $f$ represents the forecast, $c$ is the climatology (long time average measurement), and $a$ is the prediction target.

\subsection{Image Classification}
Following the approach of previous studies \cite{vit,s4nd,resnet}, \ourmethod~ is trained on the ImageNet-1K dataset for 300 epochs using the AdamW optimizer with $\beta=(0.9,0.999)$ and a learning rate of $1e-3$. We use a patch size of 8. The results are presented in \cref{tab:classification}. Our model demonstrates superior performance compared to transformer-based models when operating under similar conditions, and it achieves results on par with the state-of-the-art state-space model, S4ND \cite{vit}. We compare our results with Hyena \cite{poli2023hyena}, ViT \cite{vit}, and S4ND \cite{s4nd}. Notably, Mamba-ND-S shows a remarkable improvement of $+3.8$ in accuracy when compared to ViT, while simultaneously reducing the parameter count by 20.7\%. This performance gap is consistent with prior research on 1D sequences \cite{guMambaLinearTimeSequence2023}, where Mamba consistently outperforms transformers with fewer parameters.

\begin{table}[t]
    \centering
    \caption{\textbf{ImageNet 1K Classification Results.} We report Top 1 Accuracy on the validation set. Mamba-ND-S shows a remarkable improvement of $+3.8$ in accuracy when compared to ViT-B while reducing the parameter count to 20.7\%.}
    \begin{tabular}{l|ccc}
    Model & Image Size & Params & Acc.$\uparrow$ \\
    \specialrule{.15em}{.075em}{.075em} 
   ViT-B      & 384 & 86M & 77.9 \\
    ViT-L       & 384 & 307M & 76.5 \\
    \hline
    S4ND-ViT-B      & 224 & 86M & 80.4 \\
    Hynea-ViT-B & 224 & 88M & 78.5 \\
    \hline
    DeiT-S       & 224 & 22M & 79.8 \\
    DeiT-B       & 224 & 86M & 81.8 \\
    \hline
    Swin-T & 224 & 28M & 81.3 \\
    % Swin-S & 224 & 28M & 83.0 \\
    Swin-B & 224 & 88M & 83.5 \\
    \hline
    % Mamba1D-B   & 224 & 92M & 74.6 \\
    \rowcolor{lightgray}
    Mamba-2D-S   & 224 & 24M & 81.7 \\
    % \rowcolor{lightgray}
    % Mamba-2D-S   & 224 & 63M & 79.4 \\
    \rowcolor{lightgray}
    Mamba-2D-B   & 224 & 92M & 83.0 \\
    \end{tabular}
    \label{tab:classification}
\end{table}

\begin{table}[t]
    \centering
    \caption{\textbf{HMDB-51 and UCF-101 Video Classification Results.} All models are initialized with ImageNet weights. *: Numbers from S4ND\cite{s4nd} paper. $\dag$: Our reproduced numbers. Memory: Training memory measured in GB on a A100 GPU. All models are trained with a batch size of 16 per GPU, except S4ND, which has a batch size of 8 (OOM at 16). We also report the samples per second.}
    \begin{tabular}{l|c c c | c  c}
    Model & HDMB-51 $\uparrow$ & UCF-101 $\uparrow$ & Params & Memory & Samples/s \\
    \hline
    \textcolor{gray}{ConvNeXt-I3D*} & \textcolor{gray}{58.1} & - & \textcolor{gray}{29M} & - & - \\
    \textcolor{gray}{S4ND-ConvNeXt-3D*} & \textcolor{gray}{62.1} & - & \textcolor{gray}{29M} & - & - \\
    \hline
    Inception-I3D & 49.8 &84.5 & 25M  & - & - \\
    \hline
     ConvNeXt-I3D$\dag$ & 53.5 & 87.6 & 29M  & 17GB & 7.5 \\
     S4ND-ConvNeXt-3D$\dag$ & 56.6 & 69.3 &  29M & 77GB & 6.3\\
     \hline
    Video-Swin-T & 53.0 & 88.3 & 30M & 35GB & 22.6 \\
    Video-Swin-S & 58.1 & 88.7 & 54M  & 73GB &39.2 \\
    \hline
     \rowcolor{lightgray}
    Mamba-2D & 51.2 & 84.7 &  24M & 12GB &39.2\\
     \rowcolor{lightgray}
    Mamba-3D & 60.9 & 89.6 & 36M & 17GB & 
19.6 \\
    \end{tabular}
    \label{tab:video}
\end{table}

\begin{table}[t]
    \centering
    \label{tab:k400_results}
    \caption{\textbf{Video Classification Results on Kinetics400 and Breakfast Classification Results. } For Kinectics400, we report results using 32 frames (/32) and 64 frames (/64).}
    \subfloat[\textbf{Kinetics400 Video Classification Results.} ]{
    \centering  
    \begin{minipage}{0.3\linewidth}
        \centering
\begin{tabular}{l|Hccc}

Method & Pretrain & Views & Acc$\uparrow$ & Params \\
\hline
Swin-T & IN-1K & 4×3 & 78.8 & 28M \\
Swin-S & IN-1K & 4×3 & 80.6 & 49M \\
Swin-B & IN-1K & 4×3 & 80.6 & 88M \\
ViViT-L & ImageNet-21K & 4×3 & 80.6 & 310M \\
TimeSformer-L & ImageNet-21K & 1×3 & 80.7 & 121M \\
\hline
\rowcolor{lightgray}
Mamba-3D/32 & IN-1K & 4×3 & 80.9 & 38M \\
\rowcolor{lightgray}
Mamba-3D/64 & IN-1K & 4×3 & 81.9 & 39M \\

\end{tabular}
    \end{minipage}
    }
    \hfill
    \subfloat[\textbf{Long Video Classification Results on Breakfast dataset.}]{
    \centering  
    \begin{minipage}{0.5\linewidth}
        \centering
\begin{tabular}{l|c|c}
 & Arch. & Acc$\uparrow$  \\
\hline
Distant Supervision & TimeSformer & 89.9 \\
ViS4mer & Swin+SSM & 88.2 \\
TranS4mer & SSM & 90.3 \\
\hline
\rowcolor{lightgray}
Mamba-3D & SSM & 91.2 \\

\end{tabular}
    \end{minipage}
    }
\end{table}

\subsection{Video Action Recognition}
Prior works \cite{carreira2017quo,liu2022video} demonstrate strong performance on video datasets by adapting ImageNet-pretrained vision models to 3D tasks. Following their strategies, we inflate Mamba-2D to Mamba-3D. Since we adopted an alternating design, each layer only sees a 1D input sequence. This means the sizes of weights in Mamba-2D and Mamba-3D layers are identical, and we can directly load the Mamba-2D checkpoint. The only weight that have different shapes are the patch embeddings. We adopted a temporal patch size of 2, so we duplicated the ImageNet weights along the time dimension and divide the value by 2. To address the change in ordering from H+H-W+W- to H+H-W+W-T+T-, we simply append new layers for T+T- every other four layers.

We sample 32 frames at a frame interval of 2 from each video clip, which amounts to around 2 seconds of video. We use the AdamW optimizer with $\beta=(0.9,0.999)$ and a learning rate of $6e-4$. The learning rate on the backbone is multiplied by a factor of 0.1. We train our model with a global batch size of 64 for 50 epochs. We select Inception-I3D \cite{carreira2017quo} and ConvNeXt-I3D \cite{liu2022convnet} as our convolutional baseline. For transformer-based models, we select Video Swin Transformer \cite{liu2022video},TimeSfromerr \cite{bertasius2021space}, and ViViT\cite{arnab2021vivit}. We conducted experiments of UCF-101\cite{soomro2012ucf101}, HMDB-51\cite{kuehne2011hmdb} and Kinects-400\cite{kay2017kinetics} datasets. All models are initialized with the ImageNet-1K pretrained checkpoints. 
We report the results in \cref{tab:video} and \cref{tab:k400_results}. Our method achieves a $+1.3$ accuracy on Kinectics-400 dataset compared to Video Swin Transformer while using only 44\% of the parameters. 

To further explore the capabilities of Mamba-3D on video understanding task, we evaluate Mamba-3D on long video classification task on Breakfast\cite{kuehne2014language} dataset. Mamba-3D outperforms previous state-of-the art models based on Transformers (Distant Supervision\cite{lin2022learning}), SSMs (TranS4mer\cite{islam2023efficient}), and hybrid architectures (ViS4mer\cite{islam2022long}). 

% Compared to the state-of-the-art ConvNeXt, we achieve a $+7.4$ accuracy.

\subsection{Global Weather Forecasting}
For weather forecasting, we use Cli-ViT \cite{nguyen2023climax,nguyen2023climatelearn,nguyen2023scaling} and Pangu-Weather \cite{bi2022pangu} as our baselines. Because these models were originally trained on terabytes of high-resolution data, typically around 1$\degree$, for a long time, we cannot directly compare them due to computational constraints. Instead, we use the 5.625$\degree$ version of the ERA data across 7 pressure levels. Our baselines Cli-ViT are based on Vision Transformer, and Pangu-Weather is based on Swin-Transformer. For Cli-ViT, we use the official implementation from ClimaX \cite{nguyen2023climax}. For Pangu-Weather, we re-implement it in PyTorch. We train all models from scratch for 50 epochs, using a learning rate of 0.0005. The learning rate follows a cosine decay schedule with a warm-up period of 5 epochs. For Mamba-3D, we use a patch size of $2 \times 2 \times 2$.
We show the results in \cref{tab:era5_performance}. Compared to Cli-ViT, Mamba-3D achieved a $+0.7$ ACC while reducing the parameter count by 44.5\%.

\begin{table}[t]{
    \centering
    \caption{\textbf{ERA5 5.625$\degree$ Weather Forecasting Results.} Anomaly Correlation Coefficient (ACC) and RMSE on geopotential at the 500 hPa level. Compared to Cli-ViT, Mamba-3D achieved a $+0.7$ ACC while reducing the parameter count by 44.5\%.}
    \label{tab:era5_performance}
    \begin{tabular}{l|cccc}
       & Arch & Parms & RMSE$\downarrow$ & ACC$\uparrow$ \\
    % \specialrule{.15em}{.075em}{.075em} 
    \hline
    Cli-ViT   & ViT & 108M  & 467  & 89.3 \\
    % Pangu*     & Swin & 210M  & 464  & 89.4 \\
    Pangu     & Swin & 50M & 462  & 89.0 \\
    \hline
    \rowcolor{lightgray}
    Mamba-3D  & Mamba & 50M & 433  & 90.1 \\
    \end{tabular}
    }
\end{table}

\subsection{3D Medical Image Segmentation}
We evaluate our model on BTCV\cite{btcv} dataset. BTCV consists of abdominal CT scans of 30 subjects. Each CT scan consists of 80 to 225 slices with 512×512 pixels. We use the report numbers on split0 provided by UNETR with 6 validation samples and 24 training samples. We compare against ViT-Based method (UNETR) \cite{hatamizadeh2022unetr} and Swin-Based method (Swin-UNETR) \cite{hatamizadeh2021swin}. We keep the decoder of UNETR unchanged and only replaced its ViT backbone. We use a patch size of $16 \times 16 \times 16$. All models have 12 layers.  The Small, Small+, and Base variant of Mamba-3D-UNETR has hidden sizes of 384, 512 and 768 respectively.  We train all models on a single GPU for 5000 epochs. Memory cost are reported accordingly. We report the DICE scores. Notably, Mamba-3D-S achieves $+0.1$ gain while reducing the number of parameters by 64\% when compared with UNETR. Mamba-3D-B, which has a similar parameter count to UNETR, achieves a gain of +2.7 compared with UNETR.

\begin{table}[t]
\centering
\caption{\textbf{3D Segmentation on BTCV Dataset}. We report the score. Mamba-3D consistently outperforms baseline architectures at with less parameter count. 
% Notably, Mamba-3D-S achieves $+0.1$ gain while reducing the number of parameters by 64\% when compared with UNETR. Mamba-3D-B, which has a similar parameter count to UNETR, achieves a gain of +2.7 compared with UNETR.
}
\label{tab:my-table}
\begin{tabular}{l|c|cccc}

Arch & Arch & Params & DICE$\uparrow$ & Memory & Patch Size \\ 
\hline
UNETR & ViT & 101M & 81.4 & 10G & [16, 16, 16] \\ 
Swin-UNETR & Swin & 63M & 83.3 &21G & [2, 2, 2] \\ 
 \hline
  \rowcolor{lightgray}
 Mamba-3D-UNETR-S & Mamba & 36M & 83.1 &8G & [16, 16, 16] \\
  \rowcolor{lightgray}
Mamba-3D-UNETR-S+ & Mamba & 55M & 83.6 &9G & [16, 16, 16] \\
 \rowcolor{lightgray}
Mamba-3D-UNETR-B & Mamba & 107M & 84.7 &9G& [16, 16, 16] \\

\end{tabular}

\end{table}

\subsection{Meta Architectures}

We perform extensive ablation studies on various design choices mentioned in \cref{sec:methodology}. We show that the alternating-directional design is the simplest and most effective one among a wide range of possible choices. 

\paragraph{Layer Design}
We use ImageNet-1K and HMDB-51 classification as our benchmarks. We perform an ablation study on Mamba-2D and Mamba-3D. We adopted various designs mentioned in \cref{sec:head_dis}. Results show that our final alternating-directional design achieves stronger performance than all proposed layer-wise changes. We show results in \cref{tab:head_ablation}. Compared with the 1D-Mamba baseline, in which no special design is adopted after naively flattening the multi-dimensional sequence into a 1D sequence, we achieved a $+4.8$ accuracy on ImageNet and a $+26.9$ accuracy on HMDB-51.
Additionally, we observe that most of the proposed multi-directional designs are able to outperform the naive and bi-directional baselines by considerable margins.
% , suggesting the importance of multi-directional design.

\begin{table}[t]
    \centering
    \caption{\textbf{Ablation Study on Layer Designs.} We report top-1 accuracy on the ImageNet-1K validation set. The Alt-Directional design is the top-performing one.
}
   \begin{tabular}{l|c|cc}

   & & IN1K$\uparrow$ & HMDB-51 $\uparrow$\\
    \hline
    Alt-Directional & Block Level & 79.4 & 59.0 \\
    Multi-Head-SSM & Layer-Level & 77.6 & 51.5 \\
    ND-SSM & Layer-Level & 77.2 & 46.7 \\
    1D-SSM & - &76.4 & 34.9 \\
    Bi-SSM &  Layer-Level &74.6 & 32.1 \\

    \end{tabular}
    \label{tab:head_ablation}
\end{table}

\paragraph{Layer arrangement}
We perform comprehensive ablation studies on various organizations of SSM layers mentioned in \cref{sec:blk_dis} at the block level. In addition to three choices discussed in \cref{sec:blk_dis}, we also experiment with a hex-directional design [H+H-W+W-T+T-], in which inputs of a block are processed by multiple layers at different ordering in parallel, and the results are summed together in the end. We show the results in \cref{tab:ordering_results}. Contrary to intuition and the layer-level results, adding multi-directional design at the block level degrades the performance. We hypothesize that this behavior is caused by the reduced number of model depths. Since we keep the number of layers fixed across all designs, incorporating some level of multi-directional parallel processing will inevitably reduce the depths of the computation graph. We provide more discussion in \cref{sec:discussion}.

\label{sec:ordering_ablation}
% \begin{table}[h]
%     \centering
%     \caption{\textbf{Ablation Study on Layer Arrangement.} We report top-1 accuracy on the HMDB-51 dataset. H+H-W+W-T+T- is the top-performing design.}
%     \begin{tabular}{l|c}
%     Ordering & HMDB-51$\uparrow$ \\
%     \hline
%     H+H-W+W-T+T- & 59.0 \\
%     \text{[H+H-W+W-T+T-]} & 38.3 \\
%     \text{[H+H-][W+W-][T+T-]} & 49.8 \\
%      \text{[H+H-W+W-][T+T-]} & 47.4 \\
%     \end{tabular}
%     \label{tab:ordering_results}
% \end{table}

\begin{table}[t]
    \centering
    \caption{\textbf{Ablation Studies}}
    \subfloat[\textbf{Ablation Study on Layer Arrangement.} We report top-1 accuracy on the HMDB-51 dataset. H+H-W+W-T+T- is the top-performing design. \label{tab:ordering_results}]{
    \centering  
    \begin{minipage}{0.4\linewidth}
        \centering
        \begin{tabular}{l|c}
        Ordering & HMDB$\uparrow$ \\
        \hline
        H+H-W+W-T+T- & 59.0 \\
        \text{[H+H-W+W-T+T-]} & 38.3 \\
        \text{[H+H-][W+W-][T+T-]} & 49.8 \\
         \text{[H+H-W+W-][T+T-]} & 47.4 \\
        \end{tabular}
    \end{minipage}
    }
    \hfill
    \subfloat[\textbf{Ablation Study on Various Factorization Policies} + indicates layer factorization. In all studies, the total number of layers is fixed. 
    % We also show the training memory cost and the number of sequences.
    $B$: batch size, and $D$: length of a single dimension. \label{tab:factorization_results}]{
    \centering  
    \begin{minipage}{0.55\linewidth}
        \centering
        \begin{tabular}{l|ccc}
        Factorization & HMDB$\uparrow$ & Mem & \#Sequences \\
        \hline
        1D+1D+1D & 44.5 & 80GB & $O(BD^2)$ \\
        2D+1D & 55.8 & 77GB & $O(BD^2)$ \\
        2D+3D & 51.9 & 18GB & $O(BD)$ \\
        3D & 59.0 & 17GB  & $O(B)$ \\
        \end{tabular}
    \end{minipage}
    }
\end{table}
\begin{figure}[t]
    \centering
    \includegraphics[width=0.6\columnwidth]{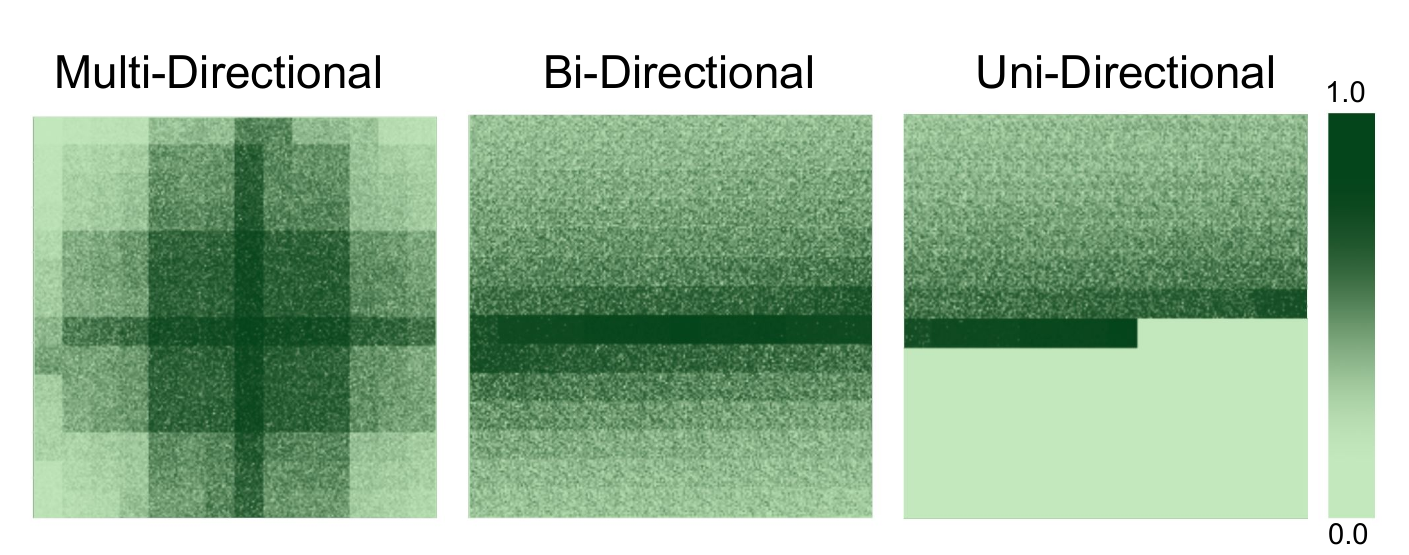}
    % \vspace{-2em}
    \caption{\textbf{Effective Receptive Field of Various Designs.} Darkness indicates the sensitivity of the central patch of the output to each pixel of the input image, normalized to the range of $(0,1)$. All images are 224x224. We use ImageNet-1K pretrained checkpoints for these visualizations.
}
    \label{fig:erf}
\end{figure}

\subsection{Scan Factorization}
While factorizing the sequence in the current implementation of Mamba leads to considerable memory and runtime overhead, these costs are external to the designs themselves and may be patched in the future. For example, a scan of N sequences of length $L$ can be considered as a long scan of one sequence of $N L$ with $\bar{A}$ from \cref{eq:1e} set to zero in places where the scan operation moves from one short sequence to another. Hence, we find it useful to study the effects of various factorization techniques despite their inferior runtime efficiency.

We show such results in \cref{tab:factorization_results}. We also report the memory cost on a single Nvidia A100 GPU. While all factorizations lead to worse performance, we find that 2D+1D factorization outperforms the 2D+3D setup, suggesting there may be merits of having certain layers dedicated to processing temporal correspondence. Because the cost of 1D factorization is high, we choose not to explore it further at this time.

\section{Discussion}
\label{sec:discussion}

\subsection{Effective Receptive Field (ERF)}
\label{sec:ablation-erf}

% To validate the significance of multi-directional design, 
We visualize the effective receptive field (ERF) \cite{luo2016understanding} of \ourmethod~ in \cref{fig:erf}. ERF is computed by setting the gradient of the center patch to one and backpropagating through the network. In particular, we visualize the ERF of ImageNet-1K pretrained weights of \ourmethod~ against the Bi-directional and Uni-directional baseline (\cref{tab:head_ablation}, row 5 and \cref{tab:head_ablation}, row 4). Sensitivity is measured by the whiteness of a pixel in the visualization. The Uni-Directional model exhibits a sharp cutoff and is insensitive to all patches after it in the flattened sequence. This is clearly undesirable in data where causal relations do not exist. The Bi-Directional model exhibits a global receptive field, but it is heavily biased in the horizontal direction. This is also undesirable because images can have vertical structures. The Multi-Directional model exhibits a more uniform pattern in the sensitivity visualization, which explains their superior performance.

\subsection{Depths versus Widths}
In \cref{sec:ordering_ablation}, we showed that a complicated multi-directional design may not necessarily lead to improved performance over the alternating-directional baseline. We hypothesize that this result is related to the effective depth of the model.

Each of the block-level designs can be represented by a directed acyclic graph (DAG) of layers, where the direction of edges implies the computation order. In this interpretation, designs other than H+H-W+W-T+T- are trading tree depths for widths. Existing literature \cite{vardi2022width} shows that depth is more important than width in deep neural networks. Our results seem to reaffirm this conclusion.

\section{Conclusion}
\label{sec:conclusion}

Transformers have been the go-to choice in image, language, and video tasks in recent years, just like ConvNets and RNNs in earlier years. Mamba \cite{mangalamReversibleVisionTransformers2022} presents itself as a competitive challenger to Transformers in 1D sequence modeling. 
% However, due to their intricate formulations discussed in \cref{sec:methodology}, it is not straightforward whether they will be equally competitive in multi-dimensional data.
In this work, we proposed \ourmethod{} which successfully extends the strong performance of Mamba to multi-dimensional inputs. \ourmethod{} outperforms Transformers on a variety of tasks with significantly fewer parameters, while incurring only subquadratic complexity. Through extensive experiments with alternative designs, we demonstrate the importance of our multi-directional design choices over uni-directional and bi-directional baselines. 
% We hope these results will facilitate future research in multi-dimensional modeling with efficient architectures.

\section{Acknowledgement}
We are grateful to Microsoft Research for supporting this research through their Accelerate Foundation Models Research program. We also thank Schmidt Sciences, Google, and Cisco for their support.

% \section*{Impact Statement}
% This paper aims to advance the field of machine learning. There are many potential social impacts of this work. However, we find no paticular ones to highlight in this section.

% \bibliography{main,refs}
% \bibliographystyle{icml2023}

%%%%%%%%%%%%%%%%%%%%%%%%%%%%%%%%%%%%%%%%%%%%%%%%%%%%%%%%%%%%%%%%%%%%%%%%%%%%%%%
%%%%%%%%%%%%%%%%%%%%%%%%%%%%%%%%%%%%%%%%%%%%%%%%%%%%%%%%%%%%%%%%%%%%%%%%%%%%%%%
% APPENDIX
%%%%%%%%%%%%%%%%%%%%%%%%%%%%%%%%%%%%%%%%%%%%%%%%%%%%%%%%%%%%%%%%%%%%%%%%%%%%%%%
%%%%%%%%%%%%%%%%%%%%%%%%%%%%%%%%%%%%%%%%%%%%%%%%%%%%%%%%%%%%%%%%%%%%%%%%%%%%%%%
% \newpage
% \appendix
% \onecolumn

% You can have as much text here as you want. The main body must be at most $8$ pages long.
% For the final version, one more page can be added.
% If you want, you can use an appendix like this one, even using the one-column format.
% %%%%%%%%%%%%%%%%%%%%%%%%%%%%%%%%%%%%%%%%%%%%%%%%%%%%%%%%%%%%%%%%%%%%%%%%%%%%%%%
% %%%%%%%%%%%%%%%%%%%%%%%%%%%%%%%%%%%%%%%%%%%%%%%%%%%%%%%%%%%%%%%%%%%%%%%%%%%%%%%

\

\bibliographystyle{splncs04}
\bibliography{egbib}

\begin{thebibliography}{10}
\providecommand{\url}[1]{\texttt{#1}}
\providecommand{\urlprefix}{URL }
\providecommand{\doi}[1]{https://doi.org/#1}

\bibitem{akbari2021vatt}
Akbari, H., Yuan, L., Qian, R., Chuang, W.H., Chang, S.F., Cui, Y., Gong, B.: Vatt: Transformers for multimodal self-supervised learning from raw video, audio and text. Advances in Neural Information Processing Systems  \textbf{34},  24206--24221 (2021)

\bibitem{arnab2021vivit}
Arnab, A., Dehghani, M., Heigold, G., Sun, C., Lu{\v{c}}i{\'c}, M., Schmid, C.: Vivit: A video vision transformer. In: Proceedings of the IEEE/CVF international conference on computer vision. pp. 6836--6846 (2021)

\bibitem{baron20232}
Baron, E., Zimerman, I., Wolf, L.: 2-d ssm: A general spatial layer for visual transformers. arXiv preprint arXiv:2306.06635  (2023)

\bibitem{bertasius2021space}
Bertasius, G., Wang, H., Torresani, L.: Is space-time attention all you need for video understanding? In: ICML. vol.~2, p.~4 (2021)

\bibitem{bi2022pangu}
Bi, K., Xie, L., Zhang, H., Chen, X., Gu, X., Tian, Q.: Pangu-weather: A 3d high-resolution model for fast and accurate global weather forecast. arXiv preprint arXiv:2211.02556  (2022)

\bibitem{carreira2017quo}
Carreira, J., Zisserman, A.: Quo vadis, action recognition? a new model and the kinetics dataset. In: proceedings of the IEEE Conference on Computer Vision and Pattern Recognition. pp. 6299--6308 (2017)

\bibitem{chen2017sca}
Chen, L., Zhang, H., Xiao, J., Nie, L., Shao, J., Liu, W., Chua, T.S.: Sca-cnn: Spatial and channel-wise attention in convolutional networks for image captioning. In: Proceedings of the IEEE conference on computer vision and pattern recognition. pp. 5659--5667 (2017)

\bibitem{christoph2016spatiotemporal}
Christoph, R., Pinz, F.A.: Spatiotemporal residual networks for video action recognition. Advances in neural information processing systems  \textbf{2} (2016)

\bibitem{dai2021coatnet}
Dai, Z., Liu, H., Le, Q.V., Tan, M.: Coatnet: Marrying convolution and attention for all data sizes. Advances in neural information processing systems  \textbf{34},  3965--3977 (2021)

\bibitem{dengImageNetLargescaleHierarchical2009}
Deng, J., Dong, W., Socher, R., Li, L.J., Li, K., {Fei-Fei}, L.: {{ImageNet}}: {{A}} large-scale hierarchical image database. In: 2009 {{IEEE Conference}} on {{Computer Vision}} and {{Pattern Recognition}}. pp. 248--255 (Jun 2009). \doi{10.1109/CVPR.2009.5206848}

\bibitem{ding2022davit}
Ding, M., Xiao, B., Codella, N., Luo, P., Wang, J., Yuan, L.: Davit: Dual attention vision transformers. In: European Conference on Computer Vision. pp. 74--92. Springer (2022)

\bibitem{dong2022cswin}
Dong, X., Bao, J., Chen, D., Zhang, W., Yu, N., Yuan, L., Chen, D., Guo, B.: Cswin transformer: A general vision transformer backbone with cross-shaped windows. In: Proceedings of the IEEE/CVF Conference on Computer Vision and Pattern Recognition. pp. 12124--12134 (2022)

\bibitem{vit}
Dosovitskiy, A., Beyer, L., Kolesnikov, A., Weissenborn, D., Zhai, X., Unterthiner, T., Dehghani, M., Minderer, M., Heigold, G., Gelly, S., et~al.: An image is worth 16x16 words: Transformers for image recognition at scale. arXiv preprint arXiv:2010.11929  (2020)

\bibitem{feichtenhofer2019slowfast}
Feichtenhofer, C., Fan, H., Malik, J., He, K.: Slowfast networks for video recognition. In: Proceedings of the IEEE/CVF international conference on computer vision. pp. 6202--6211 (2019)

\bibitem{feichtenhofer2016convolutional}
Feichtenhofer, C., Pinz, A., Zisserman, A.: Convolutional two-stream network fusion for video action recognition. In: Proceedings of the IEEE conference on computer vision and pattern recognition. pp. 1933--1941 (2016)

\bibitem{guMambaLinearTimeSequence2023}
Gu, A., Dao, T.: Mamba: {{Linear-Time Sequence Modeling}} with {{Selective State Spaces}} (Dec 2023). \doi{10.48550/arXiv.2312.00752}

\bibitem{s4}
Gu, A., Goel, K., R\'e, C.: Efficiently modeling long sequences with structured state spaces. In: The International Conference on Learning Representations ({ICLR}) (2022)

\bibitem{hatamizadeh2021swin}
Hatamizadeh, A., Nath, V., Tang, Y., Yang, D., Roth, H.R., Xu, D.: Swin unetr: Swin transformers for semantic segmentation of brain tumors in mri images. In: International MICCAI Brainlesion Workshop. pp. 272--284. Springer (2021)

\bibitem{hatamizadeh2022unetr}
Hatamizadeh, A., Tang, Y., Nath, V., Yang, D., Myronenko, A., Landman, B., Roth, H.R., Xu, D.: Unetr: Transformers for 3d medical image segmentation. In: Proceedings of the IEEE/CVF winter conference on applications of computer vision. pp. 574--584 (2022)

\bibitem{resnet}
He, K., Zhang, X., Ren, S., Sun, J.: Deep residual learning for image recognition. In: Proceedings of the IEEE conference on computer vision and pattern recognition. pp. 770--778 (2016)

\bibitem{hersbach2018era5}
Hersbach, H., Bell, B., Berrisford, P., Biavati, G., Hor{\'a}nyi, A., Mu{\~n}oz~Sabater, J., Nicolas, J., Peubey, C., Radu, R., Rozum, I., et~al.: Era5 hourly data on single levels from 1979 to present. Copernicus climate change service (c3s) climate data store (cds)  \textbf{10}(10.24381) (2018)

\bibitem{ho2019axial}
Ho, J., Kalchbrenner, N., Weissenborn, D., Salimans, T.: Axial attention in multidimensional transformers. arXiv preprint arXiv:1912.12180  (2019)

\bibitem{huang2017densely}
Huang, G., Liu, Z., Van Der~Maaten, L., Weinberger, K.Q.: Densely connected convolutional networks. In: Proceedings of the IEEE conference on computer vision and pattern recognition. pp. 4700--4708 (2017)

\bibitem{islam2022long}
Islam, M.M., Bertasius, G.: Long movie clip classification with state-space video models. arXiv preprint arXiv:2204.01692  (2022)

\bibitem{islam2023efficient}
Islam, M.M., Hasan, M., Athrey, K.S., Braskich, T., Bertasius, G.: Efficient movie scene detection using state-space transformers. In: Proceedings of the IEEE/CVF Conference on Computer Vision and Pattern Recognition. pp. 18749--18758 (2023)

\bibitem{kay2017kinetics}
Kay, W., Carreira, J., Simonyan, K., Zhang, B., Hillier, C., Vijayanarasimhan, S., Viola, F., Green, T., Back, T., Natsev, P., et~al.: The kinetics human action video dataset. arXiv preprint arXiv:1705.06950  (2017)

\bibitem{krizhevsky2012imagenet}
Krizhevsky, A., Sutskever, I., Hinton, G.E.: Imagenet classification with deep convolutional neural networks. Advances in neural information processing systems  \textbf{25} (2012)

\bibitem{kuehne2014language}
Kuehne, H., Arslan, A., Serre, T.: The language of actions: Recovering the syntax and semantics of goal-directed human activities. In: Proceedings of the IEEE conference on computer vision and pattern recognition. pp. 780--787 (2014)

\bibitem{kuehne2011hmdb}
Kuehne, H., Jhuang, H., Garrote, E., Poggio, T., Serre, T.: Hmdb: a large video database for human motion recognition. In: 2011 International conference on computer vision. pp. 2556--2563. IEEE (2011)

\bibitem{btcv}
Landman, B., Xu, Z., Igelsias, J., Styner, M., Langerak, T., Klein, A.: Miccai multi-atlas labeling beyond the cranial vault--workshop and challenge. In: Proc. MICCAI Multi-Atlas Labeling Beyond Cranial Vault—Workshop Challenge. vol.~5, p.~12 (2015)

\bibitem{lecunHandwrittenDigitRecognition1989}
LeCun, Y., Boser, B., Denker, J., Henderson, D., Howard, R., Hubbard, W., Jackel, L.: Handwritten {{Digit Recognition}} with a {{Back-Propagation Network}}. In: Advances in {{Neural Information Processing Systems}}. vol.~2. {Morgan-Kaufmann} (1989)

\bibitem{li2020attention}
Li, J., Jin, K., Zhou, D., Kubota, N., Ju, Z.: Attention mechanism-based cnn for facial expression recognition. Neurocomputing  \textbf{411},  340--350 (2020)

\bibitem{li2022mvitv2}
Li, Y., Wu, C.Y., Fan, H., Mangalam, K., Xiong, B., Malik, J., Feichtenhofer, C.: Mvitv2: Improved multiscale vision transformers for classification and detection. In: Proceedings of the IEEE/CVF Conference on Computer Vision and Pattern Recognition. pp. 4804--4814 (2022)

\bibitem{lin2022learning}
Lin, X., Petroni, F., Bertasius, G., Rohrbach, M., Chang, S.F., Torresani, L.: Learning to recognize procedural activities with distant supervision. In: Proceedings of the IEEE/CVF Conference on Computer Vision and Pattern Recognition. pp. 13853--13863 (2022)

\bibitem{liu2024vmamba}
Liu, Y., Tian, Y., Zhao, Y., Yu, H., Xie, L., Wang, Y., Ye, Q., Liu, Y.: Vmamba: Visual state space model. arXiv preprint arXiv:2401.10166  (2024)

\bibitem{liu2021swin}
Liu, Z., Lin, Y., Cao, Y., Hu, H., Wei, Y., Zhang, Z., Lin, S., Guo, B.: Swin transformer: Hierarchical vision transformer using shifted windows. In: Proceedings of the IEEE/CVF international conference on computer vision. pp. 10012--10022 (2021)

\bibitem{liu2022video}
Liu, Z., Ning, J., Cao, Y., Wei, Y., Zhang, Z., Lin, S., Hu, H.: Video swin transformer. In: Proceedings of the IEEE/CVF conference on computer vision and pattern recognition. pp. 3202--3211 (2022)

\bibitem{liu2022convnet}
Liu, Z., Mao, H., Wu, C.Y., Feichtenhofer, C., Darrell, T., Xie, S.: A convnet for the 2020s. In: Proceedings of the IEEE/CVF conference on computer vision and pattern recognition. pp. 11976--11986 (2022)

\bibitem{lu2021container}
Lu, J., Mottaghi, R., Kembhavi, A., et~al.: Container: Context aggregation networks. Advances in neural information processing systems  \textbf{34},  19160--19171 (2021)

\bibitem{luo2016understanding}
Luo, W., Li, Y., Urtasun, R., Zemel, R.: Understanding the effective receptive field in deep convolutional neural networks. Advances in neural information processing systems  \textbf{29} (2016)

\bibitem{U-Mamba}
Ma, J., Li, F., Wang, B.: U-mamba: Enhancing long-range dependency for biomedical image segmentation. arXiv preprint arXiv:2401.04722  (2024)

\bibitem{mangalamReversibleVisionTransformers2022}
Mangalam, K., Fan, H., Li, Y., Wu, C.Y., Xiong, B., Feichtenhofer, C., Malik, J.: Reversible {{Vision Transformers}}. In: Proceedings of the {{IEEE}}/{{CVF Conference}} on {{Computer Vision}} and {{Pattern Recognition}}. pp. 10830--10840 (2022)

\bibitem{neimark2021video}
Neimark, D., Bar, O., Zohar, M., Asselmann, D.: Video transformer network. In: Proceedings of the IEEE/CVF international conference on computer vision. pp. 3163--3172 (2021)

\bibitem{s4nd}
Nguyen, E., Goel, K., Gu, A., Downs, G.W., Shah, P., Dao, T., Baccus, S.A., R\'e, C.: S4nd: Modeling images and videos as multidimensional signals using state spaces. Advances in Neural Information Processing Systems  \textbf{35} (2022)

\bibitem{nguyen2023climax}
Nguyen, T., Brandstetter, J., Kapoor, A., Gupta, J.K., Grover, A.: Climax: A foundation model for weather and climate. arXiv preprint arXiv:2301.10343  (2023)

\bibitem{nguyen2023climatelearn}
Nguyen, T., Jewik, J., Bansal, H., Sharma, P., Grover, A.: Climatelearn: Benchmarking machine learning for weather and climate modeling. arXiv preprint arXiv:2307.01909  (2023)

\bibitem{nguyen2023scaling}
Nguyen, T., Shah, R., Bansal, H., Arcomano, T., Madireddy, S., Maulik, R., Kotamarthi, V., Foster, I., Grover, A.: Scaling transformer neural networks for skillful and reliable medium-range weather forecasting. arXiv preprint arXiv:2312.03876  (2023)

\bibitem{poli2023hyena}
Poli, M., Massaroli, S., Nguyen, E., Fu, D.Y., Dao, T., Baccus, S., Bengio, Y., Ermon, S., R{\'e}, C.: Hyena hierarchy: Towards larger convolutional language models. arXiv preprint arXiv:2302.10866  (2023)

\bibitem{simonyan2014very}
Simonyan, K., Zisserman, A.: Very deep convolutional networks for large-scale image recognition. arXiv preprint arXiv:1409.1556  (2014)

\bibitem{soomro2012ucf101}
Soomro, K., Zamir, A.R., Shah, M.: Ucf101: A dataset of 101 human actions classes from videos in the wild. arXiv preprint arXiv:1212.0402  (2012)

\bibitem{szegedy2015going}
Szegedy, C., Liu, W., Jia, Y., Sermanet, P., Reed, S., Anguelov, D., Erhan, D., Vanhoucke, V., Rabinovich, A.: Going deeper with convolutions. In: Proceedings of the IEEE conference on computer vision and pattern recognition. pp.~1--9 (2015)

\bibitem{tian2023integrally}
Tian, Y., Xie, L., Wang, Z., Wei, L., Zhang, X., Jiao, J., Wang, Y., Tian, Q., Ye, Q.: Integrally pre-trained transformer pyramid networks. In: Proceedings of the IEEE/CVF Conference on Computer Vision and Pattern Recognition. pp. 18610--18620 (2023)

\bibitem{tran2015learning}
Tran, D., Bourdev, L., Fergus, R., Torresani, L., Paluri, M.: Learning spatiotemporal features with 3d convolutional networks. In: Proceedings of the IEEE international conference on computer vision. pp. 4489--4497 (2015)

\bibitem{tran2018closer}
Tran, D., Wang, H., Torresani, L., Ray, J., LeCun, Y., Paluri, M.: A closer look at spatiotemporal convolutions for action recognition. In: Proceedings of the IEEE conference on Computer Vision and Pattern Recognition. pp. 6450--6459 (2018)

\bibitem{van2016pixel}
Van Den~Oord, A., Kalchbrenner, N., Kavukcuoglu, K.: Pixel recurrent neural networks. In: International conference on machine learning. pp. 1747--1756. PMLR (2016)

\bibitem{vardi2022width}
Vardi, G., Yehudai, G., Shamir, O.: Width is less important than depth in relu neural networks. In: Conference on Learning Theory. pp. 1249--1281. PMLR (2022)

\bibitem{vaswaniAttentionAllYou2017}
Vaswani, A., Shazeer, N., Parmar, N., Uszkoreit, J., Jones, L., Gomez, A.N., Kaiser, {\L}., Polosukhin, I.: Attention is {{All}} you {{Need}}. In: Advances in {{Neural Information Processing Systems}}. vol.~30. {Curran Associates, Inc.} (2017)

\bibitem{wang2021pyramid}
Wang, W., Xie, E., Li, X., Fan, D.P., Song, K., Liang, D., Lu, T., Luo, P., Shao, L.: Pyramid vision transformer: A versatile backbone for dense prediction without convolutions. In: Proceedings of the IEEE/CVF international conference on computer vision. pp. 568--578 (2021)

\bibitem{wang2018non}
Wang, X., Girshick, R., Gupta, A., He, K.: Non-local neural networks. In: Proceedings of the IEEE conference on computer vision and pattern recognition. pp. 7794--7803 (2018)

\bibitem{zhang2016real}
Zhang, B., Wang, L., Wang, Z., Qiao, Y., Wang, H.: Real-time action recognition with enhanced motion vector cnns. In: Proceedings of the IEEE conference on computer vision and pattern recognition. pp. 2718--2726 (2016)

\bibitem{zhang2018real}
Zhang, B., Wang, L., Wang, Z., Qiao, Y., Wang, H.: Real-time action recognition with deeply transferred motion vector cnns. IEEE Transactions on Image Processing  \textbf{27}(5),  2326--2339 (2018)

\bibitem{zhang2022hivit}
Zhang, X., Tian, Y., Xie, L., Huang, W., Dai, Q., Ye, Q., Tian, Q.: Hivit: A simpler and more efficient design of hierarchical vision transformer. In: The Eleventh International Conference on Learning Representations (2022)

\bibitem{zhu2024vision}
Zhu, L., Liao, B., Zhang, Q., Wang, X., Liu, W., Wang, X.: Vision mamba: Efficient visual representation learning with bidirectional state space model (2024)

\end{thebibliography}

\newpage
\maketitlesupplementary{}

% \subsection{Erratas}
% We correct the following differences of model performance in the main paper: on HMDB-51, our method achieves a +2.8 (instead of +0.9) accuracy compared to Video Swin Transformer  and +7.4 (instead of +5.5) accuracy compared to ConvNeXT. We also correct the learning rate to be 6.0e-4 (instead of 1.0e-3) for the main results. For completeness, we provide results under different learnig rate in \cref{tab:learning_rate}. While learning rate affects model performance by a non-trival margin, our model consistently outperforms competitors regardless of learning rates. 

% \begin{table}[h]
%     \centering
%     \caption{\textbf{Ablation Studies on Learning Rate}}
%     \textbf{Ablation Study on Learning Rate.} We report top-1 accuracy on the HMDB-51 dataset of swpeeing learning rate between 1.0e-3 and 1.0e-4.  \label{tab:learning_rate}{
%     \centering  
%     \begin{minipage}{0.4\linewidth}
%         \centering
%         \begin{tabular}{l|c}
%         Learning rate & HMDB-51$\uparrow$ \\
%         \hline
%         1.0e-3 & 59.0 \\
%         6.0e-4 & 60.9 \\
%         1.0e-4 & 60.2 \\
%         \end{tabular}
%     \end{minipage}
%     }
% \end{table}

\section{Additional Ablation Studies}

\subsection{Additional ImageNet and HMDB-51 Results.}
HMDB-51 and UCF-101 training requires ImageNet pretrained weights. Due to architectural differences, different HMDB-51 experiments may load different ImageNet pretrained weights. For completeness, we provide full results indicating which weights are loaded for each experiment. All models are trained with a learning rate of 1e-3 and a patch size of 16 for both pretraining and finetuning. We report the results in \cref{tab:ablation_in1k_full}.

In addition to the results already referred to in the main paper, we also incorporate additional experiments on applying an alternating design to Bi-SSM. In particular, the default Bi-SSM design processes each layer in $L+$ and $L-$ direction, which is equivalent to $H+$ and $H-$. We experiment with processing the input in $H+$ and $H-$ direction in odd layers and in $W+$ and $W-$ direction in even layers. This leads to a considerable improvement in both ImageNet-1K (+1.5) and HMDB-51 (+8.4), highlighting the effectiveness of alternating directional designs.

We also tried to train models from scratch on HMDB-51. These models all converge very slowly or diverge. We failed to obtain useful results.

\begin{table}[h]
\centering
\caption{\textbf{Detailed Ablation Results of various combinations of design choices.} In the 2D Layer column, we list the layer-level design for ImageNet pretraining. In the 3D Layer column, we list the layer-level design for finetuning. (.) denotes layer-level grouping. For Bi-SSM, each layer incorporates both the forward direction and reverse direction, hence no $+/-$ distinction is needed.
}
\label{tab:ablation_in1k_full}
\begin{tabular}{c|c|c|c|c}

Kernel & 2D Layer & IN1K$\uparrow$ & 3D Layer & HMDB-51$\uparrow$  \\
\hline
1D-SSM                   & L+                     & 76.4                       & L+                         & 34.9                        \\ \hline
1D-SSM                   & (H+H-)(W+W-)           & 76.5                       & (H+H-)(W+W-)(T+T-)         & 49.8                        \\ \hline
\multirow{2}{*}{1D-SSM}                  & \multirow{2}{*}{H+H-W+W- }              & \multirow{2}{*}{\textbf{79.4}}                       & (H+H-W+W-)(T+T-)           & 47.4                        \\ \cline{4-5}
                         &                        &                            & H+H-W+W-T+T-               & \textbf{59.0 }                         \\ \hline
1D-SSM                   & (L+L-)                 & 76.3                       & (L+L-)                     & 46.3                        \\ \hline
Bi-SSM*                   & L                      & 74.6                       & L                          & 32.1                        \\ \hline
Bi-SSM*                   & HW                   & 76.1                       & HWT                        & 40.5                        \\ \hline
ND-SSM                   & -                      & 77.2                       & -                          & 46.7                        \\ \hline
Multi-Head               & -                      & 77.6                       & -                          & 51.5                        \\ 
\end{tabular}

\end{table}

\subsection{Video-Specific Ablations}

\subsection{$\Delta$ Initialization}
When initializing new SSM layers in T+T- direction, we can adjust the scale of $\Delta$ to control the initial temporal receptive field. We show such results in \cref{tab:delta_scale_results}. We find that 1.0 is the optimal value, which coincides with the prior conclusion from S4ND \cite{s4nd}.

\subsection{Weight Inflation}
There are various alternatives for inflating the positional embedding from 2D to 3D. We considered two policies. Given a 3D sequence of shape $H\times W \times T$, the first option copies the original 2D embedding of shape $H \times W$ by $T$ times and scales the values to $\frac{1}{T}$. The second option copies the 2D embedding once and places it at location $\frac{T}{2}$ on the temporal axis. All other embeddings are initialized to 0. We find no significant differences in performance between these two designs.

\begin{table}[h]
    \centering
    \caption{\textbf{Ablation Study on the Scaling Factor of $\Delta$ for HMDB-51 Experiments}. We report the performance on HMDB-51 under different initializations of temporal layers (T+ and T-). $\Delta=1.0$ is the optimal choice.}
    \begin{tabular}{l|c}
    $\Delta$  & HMDB-51$\uparrow$ \\
        \hline
    0.1 & 58.0 \\
    0.2 & 57.8 \\
    1.0 & 59.0 \\
    5.0 & 55.3 \\
    \end{tabular}
    \label{tab:delta_scale_results}
\end{table}

\subsection{Ablation on 3D Segmentation}
Because we train 3D segmentation models from scratch instead of from 2D-pretrained initializations, it may offer better insight into the design choices in the 3D space. However, it may also be the case that segmentation is local in nature and may require less global information. For completeness, we compare the performance of different SSM designs in \cref{tab:head_ablation_3d}. Results show that our simple design of alternating the scan direction between layers is still the best performing one. However, 1D-SSM with no bidirectional design also exhibits strong performance.

\begin{table}[h]
    \centering
    \caption{\textbf{Ablation Study on Layer Designs.} We report top-1 accuracy on the ImageNet-1K validation set and HMDB-51 split 1. We report Dice score on BTCV dataset. The Alt-Directional design is the top-performing one.
}
   \begin{tabular}{l|c|ccc}

   & & IN1K$\uparrow$ & HMDB-51 $\uparrow$ & BTCV $\uparrow$\\
    \hline
    Alt-Directional & Block Level & 79.4 & 59.0 & 81.5  \\
    Multi-Head & Layer-Level & 77.6 & 51.5  & 81.4\\
    ND-SSM & Layer-Level & 77.2 & 46.7 & 80.2\\
    1D-SSM & - &76.4 & 34.9 & 80.7\\
    Bi-SSM &  Layer-Level &74.6 & 32.1 & 77.9 \\

    \end{tabular}
    \label{tab:head_ablation_3d}
\end{table}

\subsection{Patch Size}

\begin{figure}[h]
    \centering
    \includegraphics[width=0.6\columnwidth]{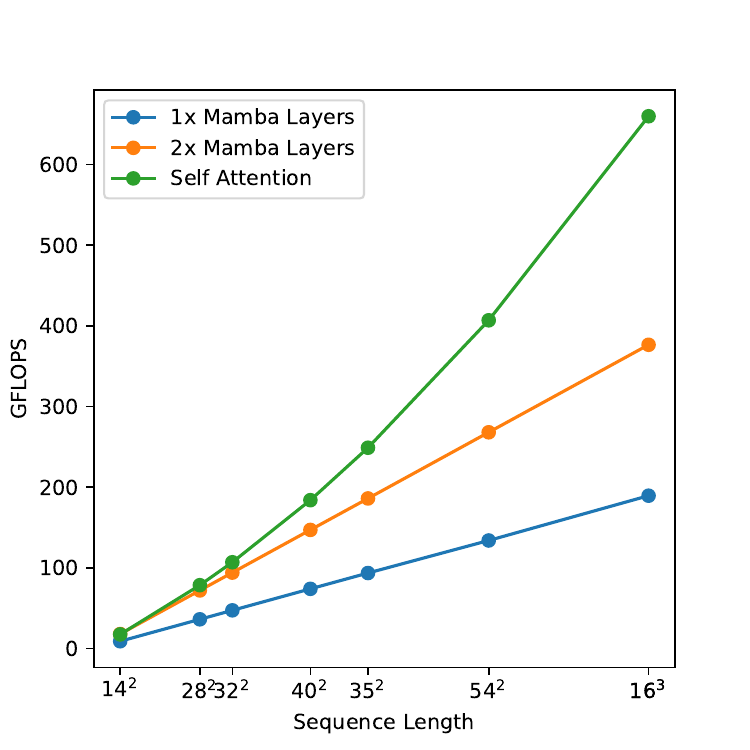}
    % \vspace{-2em}
    \caption{\textbf{\ourmethod~ Computational Complexity with Respect to Sequence Length}. We plot the computational cost of the standard 12-layer Vision Transformer (ViT-B), and comparable Mamba-2D with 12 and 24 Mamba layers. In terms of parameter count, 2 Mamba layers are roughly equivalent to 1 ViT layer, which includes a Multi-head Attention (MHA) module and a Feed Forward Network (FFN). $14^2$ reflects the standard $224 \times 224$ image with a patch size of 16. Sequence length increase can either reflect a lower patch size or a higher input resolution.}
    \label{fig:complexity}
\end{figure}

One crucial element of Mamba-ND's success depends on its capability to achieve a global context at linear complexity. This allows it to use a lower patch size or higher resolution input given fixed computation capacity when compared with self attention based methods. We highlight this in figure \cref{fig:complexity}.  On ImageNet-1K, we explore the effect of different patch sizes. We report the results in \cref{tab:patch_size}

\begin{table}[h]
    \centering
    \caption{\textbf{Ablation Study on patch size}. We report the performance on ImageNet-1K. We find that lower patch size achieves better result.
}
    \begin{tabular}{c|c c}
      &patch size  & IN1K$\uparrow$ \\
      \hline
       ViT-B & 16 & 77.9 \\
       DeiT-S & 16 & 79.8 \\
       \hline
        DeiT-S & 8 & 75.2 \\
        \hline
       Swin-T & 4 & 81.3 \\
        \hline
\rowcolor{lightgray}        
    Mamba-2D-S & 8 & 81.7 \\
    \rowcolor{lightgray}
    Mamba-2D-S & 16 & 79.4 \\
    \end{tabular}
    \label{tab:patch_size}
\end{table}

\subsection{Ordering}

In the alternating directional setup, there is still room for exploration in the ordering of axes. For example, there is no reason to assume TWH+ is better than WTH+ for H+. There is also no intuitive reason to prefer H+ over W+ as the direction of the first layer. This means the design space incorporates $2^3 \cdot 3!=48$ possible choices (there are 2 options for each of H, W, T, and there are 3! orderings of the first three layers). Due to computational constraints, we are unable to explore all options. We attempted two incremental changes from the base setup on BTCV dataset: 1) Changing the order of the first three layers to WHT and TWH. 2) Swapping the order of TWH+ to WTH+. We found no meaningful difference between the results. We hypothesize that the residual connection is sufficient to mitigate the effect of different orderings.

\section{Implementation Details}
\subsection{Model Size}
Mamba-2D-S has 24 layers and a hidden size of 384. Mamba-2D-B has 24 layers and a hidden size of 768. Mamba-3D-S has 32 layers and a hidden size of 384. Mamba-3D-S+ has 32 layers and a hidden size of 512. Mamba-3D-B has 32 layers and a hidden size of 768. For ERA5 weather forecast, we use a 12 layer version of Mamba-3D-B. In terms of the parameter count, two Mamba Layers is roughly equivalent to one ViT block of the same dimension.

\subsection{Additional Hyperparameters}
On HMDB-51, we use a learning rate of 6.0e-4. On UCF-101, we use a learning rate of 1.0e-3. We set RandAug to $(9,4)$ for both experiments. We do not use label smoothing or cutmix.  We set the dropout and drop path rate both to 0.1. On ImageNet-1K, we use a learning rate of 1.0e-3. We set RandAug to (2, 10).  We set label smoothing to 0.1, Mixup to 0.8 and CutMix to 1.0. We set the dropout and drop path rate both to 0.1. For all image and video ablations, we fix the learning rate to be 1e-3 and use a patch size of 16. On BTCV, we use a learning rate of 1.0e-4 and a drop rate of  0.1. On ERA5, we use a learning rate of 5.0e-4 and a drop rate of  0.1.

\section{Limitations}
While we have provided a relatively comprehensive set of experiments to explore the design space of multi-dimensional selective state-space models, there remain unexplored areas due to the exponential number of possible orderings. In this paper, we primarily explored various combinations of row-major ordering. However, there may be other reasonable choices such as diagonal or zig-zag patterns. We leave these possibilities for future work.

\section{Concurrent Works}
There have been recent works such as VisionMamba \cite{zhu2024vision}, VMamba \cite{liu2024vmamba}, and U-Mamba \cite{U-Mamba} that focus on applying Mamba to specific application fields such as classification or segmentation. Contrary to these works, which have a limited design space, \ourmethod~ explores the generic recipe to adopt Mamba-Kernel to multi-dimensional data, which is under-explored. For example, VisionMamba can be considered as a variant of Bi-SSM with additional designs, and VMamba can be considered as hierarchical ND-SSM with additional designs.

\end{document}